%% file: access.tex
\documentclass{ieeeaccess}
\usepackage{cite}
\usepackage{amsmath,amssymb,amsfonts}
\usepackage{graphicx}
\usepackage{textcomp}

\usepackage[caption=false]{subfig}

\usepackage{algorithm}
\usepackage[algo2e,ruled,linesnumbered]{algorithm2e}

\makeatletter
\newcommand*\bigcdot{\mathpalette\bigcdot@{.5}}
\newcommand*\bigcdot@[2]{\mathbin{\vcenter{\hbox{\scalebox{#2}{$\m@th#1\bullet$}}}}}

\newcommand{\algorithmfootnote}[2][\footnotesize]{%
  \let\old@algocf@finish\@algocf@finish
  \def\@algocf@finish{\old@algocf@finish
    \leavevmode\rlap{\begin{minipage}{\linewidth}
    #1#2
    \end{minipage}}%
  }%
}

\newcommand{\removelatexerror}{\let\@latex@error\@gobble}
\makeatother

\def\BibTeX{{\rm B\kern-.05em{\sc i\kern-.025em b}\kern-.08em
    T\kern-.1667em\lower.7ex\hbox{E}\kern-.125emX}}
\begin{document}
\doi{}

\title{A Deep Learning Approach for Predicting Spatiotemporal Dynamics from Sparsely Observed Data}

\author{\uppercase{Priyabrata Saha}, \IEEEmembership{Graduate Student Member, IEEE},
\uppercase{and Saibal Mukhopadhyay},
\IEEEmembership{Fellow, IEEE}}
\address{School  of  Electrical  and  Computer  Engineering, Georgia Institute of Technology, Atlanta, GA 30332 USA (e-mails: priyabratasaha@gatech.edu, saibal@ece.gatech.edu)}

\tfootnote{This work is supported in part by the Office of Naval Research under Grant N00014-19-1-2639.}

\markboth
{Saha \headeretal: A Deep Learning Approach for Predicting Spatiotemporal Dynamics from Sparsely Observed Data}
{Saha \headeretal: A Deep Learning Approach for Predicting Spatiotemporal Dynamics from Sparsely Observed Data}

\corresp{Corresponding author: Saibal Mukhopadhyay (e-mail: saibal@ece.gatech.edu).}

\begin{abstract}
In this paper, we consider the problem of learning prediction models for spatiotemporal physical processes driven by unknown partial differential equations (PDEs). We propose a deep learning framework that learns the underlying dynamics and predicts its evolution using sparsely distributed data sites. Deep learning has shown promising results in modeling physical dynamics in recent years. However, most of the existing deep learning methods for modeling physical dynamics either focus on solving known PDEs or require data in a dense grid when the governing PDEs are unknown. In contrast, our method focuses on learning prediction models for unknown PDE-driven dynamics only from sparsely observed data. The proposed method is spatial dimension-independent and geometrically flexible. We demonstrate our method in the forecasting task for the two-dimensional wave equation and the Burgers-Fisher equation in multiple geometries with different boundary conditions, and the ten-dimensional heat equation. 
\end{abstract}


\titlepgskip=-15pt

\maketitle

\input{introduction}
\input{backgroud}

\input{approach}
\input{experiments}
\input{conclusion}

\section*{Acknowledgment}
This work is supported in part by the Office of Naval Research under Grant N00014-19-1-2639. The views and conclusions contained in this document are those of the authors and should not be interpreted as representing the official policies, either expressed or implied, of the the Office of Naval Research or the U.S. Government.


\bibliographystyle{IEEEtran}
\bibliography{ref}

\begin{IEEEbiography}[{\includegraphics[width=1in,height=1.25in,clip,keepaspectratio]{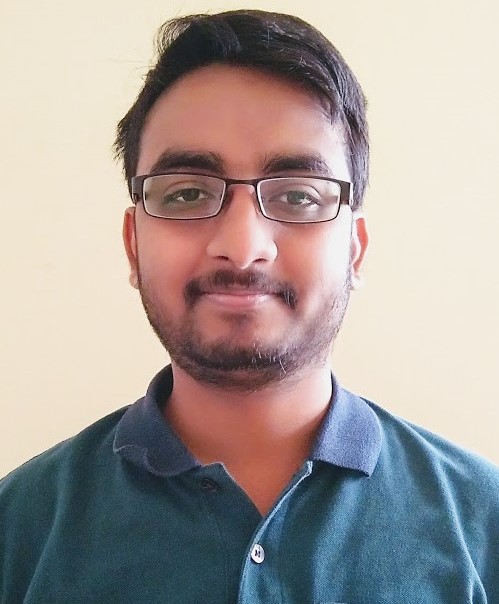}}]{Priyabrata Saha} (S'19) received the B.Tech. and M.Tech. degrees in electronics and electrical communication engineering from IIT Kharagpur, Kharagpur, India, in 2015. He is currently pursuing the Ph.D. degree in electrical and computer engineering with the Georgia Institute of Technology, Atlanta, GA, USA, under the supervision of Prof. S. Mukhopadhyay. His current research interests include multimodal computer vision, machine learning for dynamics and control, machine learning for physical sciences.
\end{IEEEbiography}

\begin{IEEEbiography}[{\includegraphics[width=1in,height=1.25in,clip,keepaspectratio]{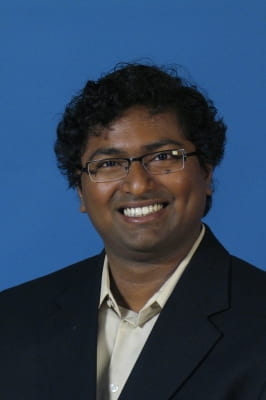}}]{DR. SAIBAL MUKHOPADHYAY} (S’99 – M’07
– SM’11 – F’18) received the B.E. degree in electronics
and telecommunication engineering from Jadavpur University, Kolkata, India, and the Ph.D. degree in electrical and computer engineering from Purdue University, West Lafayette, IN, in 2000
and 2006, respectively. Dr. Mukhopadhyay was a Research Staff Member in IBM T. J. Watson Research Center, Yorktown Heights, NY from August 2007 to September 2007. He is currently a Joseph M. Pettit Professor with the School of Electrical and Computer Engineering, Georgia Institute of Technology, Atlanta. His research interests include design of energy-efficient, intelligent, and secure systems in nanometer technologies. Dr. Mukhopadhyay was a recipient of the Office of Naval Research Young Investigator Award in 2012, the National Science Foundation CAREER Award in 2011, the IBM Faculty Partnership Award in 2009 and 2010, the SRC Inventor Recognition Award in 2008, the SRC Technical Excellence Award in 2005, the IBM Ph.D. Fellowship Award for years 2004 to 2005. He has authored or co-authored over 200 papers in refereed journals and conferences, and holds five U.S. patents. Dr. Mukhopadhyay is a Fellow of IEEE.
\end{IEEEbiography}

\EOD

\end{document}

%% file: introduction.tex
\section{Introduction}

\begin{figure*}[t]
    \centering
    \includegraphics[width=1\linewidth]{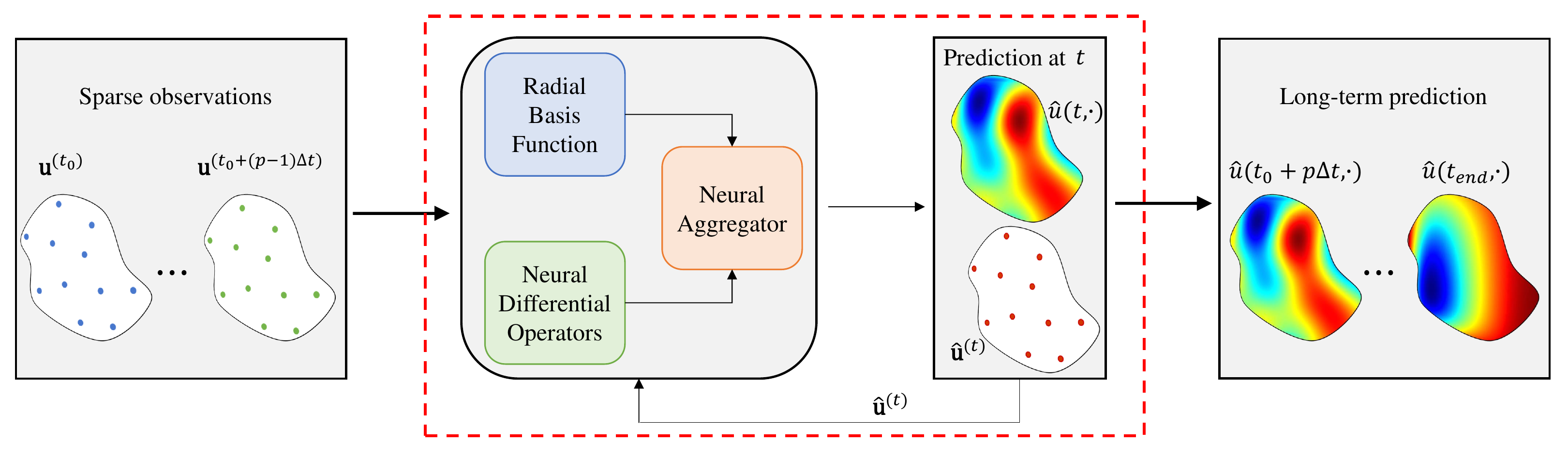}
    \caption{Iterative prediction using the learned model. Initial sparse observations are used as the input for the very first prediction. In subsequent steps, the current sparse prediction is iteratively used as the input to model for long-term prediction. $\mathbf{u}^{(t)} = [u(t, \mathbf{x}_1), \cdots, u(t, \mathbf{x}_N)]^T$ denotes the sparse measurement vector at time $t$.} 
    \label{fig:intro}
\end{figure*}

Complex physical phenomena are often modeled with a set of differential equations (DEs). Traditionally, model equations are constructed based on physical laws, rigorous formalization, and experiments by human experts. However, in many practical scenarios, analytical modeling is very challenging and/or can only describe the system behavior partially. Data-driven modeling is required in such cases to solve the forward and inverse problems. The success of deep learning in various sequence prediction tasks, for example, natural language and speech modeling \cite{sutskever2014sequence, vaswani2017attention,chung2014empirical}, video prediction \cite{xingjian2015convolutional, finn2016unsupervised, lee2017desire}, has motivated many researchers to exploit deep learning for automatic modeling of physical processes from measured or observed data. Consequently, a number of deep learning methods have been introduced for data-driven modeling of complex dynamical systems involving partial differential equations (PDEs) \cite{long2018pde, long2018hybridnet, de2019deep, saha2020phicnet, geneva2020modeling}. Apart from deep learning, other machine learning techniques, e.g. sparse optimization, are also being used to discover dynamics from data \cite{rudy2017data,champion2020unified}. Typically, deep learning methods require a significant amount of data to obtain generalized solutions and assume densely sampled system states are available for learning. However, in many practical scenarios, such a huge amount of densely sampled data are very difficult to obtain if not impossible. For example, sensors can be placed only at a few locations for data collection. This is a common situation in many fields of applied sciences where spatiotemporal prediction of state evolution is necessary.

In this paper, we consider the problem of learning prediction models for unknown PDE-driven spatiotemporal dynamics from sparsely-observed data. Formally, given a set of $N$ distinct scattered spatial locations $X = \{\mathbf{x}_1, \cdots, \mathbf{x}_N \} \subset \Omega \subset \mathbb{R}^d$ and corresponding time series of measurements of a physical quantity at those locations $\{u(t, \mathbf{x}_i) \in \mathbb{R} : \mathbf{x}_i \in X \ \text{and} \  t=t_0, t_0 + \Delta t,  t_0 + 2 \Delta t, \cdots\}$, we want to learn a deep neural network model that can predict the spatiotemporal evolution of the underlying physical process. We assume that the system dynamics is governed by a PDE belonging to a family of PDEs described by the following generic form:
\begin{equation}
    \partial_t^p u = F(u, \partial_{\mathbf{x}} u,  \partial_{\mathbf{x}}^2 u, \cdots), \quad \mathbf{x} \in \Omega \subset \mathbb{R}^d, \quad t > 0.
    \label{eqn:problem}
\end{equation}
$\partial^p_t$ denotes the $p^{th}$ order partial differential operator with respect to time and $\partial^q_\mathbf{x}$ denotes all possible combinations of $q^{th}$ order spatial (partial) differential operators. We assume that the temporal order ($p$) of the underlying PDE is known \textit{a priori} while the form of the function $F$, nonlinear in general, and its arguments are unknown. Though we observe the system at few discrete locations, we seek to predict the time evolution of $u$ at any generic location $\mathbf{x} \in \Omega$ or over the entire $\Omega$. We want the learned model to forecast the evolution of the system for any given initial and boundary conditions (same or different from the training conditions) provided that the underlying PDE remains the same. Figure \ref{fig:intro} shows the process of iterative prediction using a learned model. While the model provides prediction at any generic location of the domain, we only use the prediction at the measurement locations as the input for the next step to keep the computational expenses at a minimum. 

Our approach integrates the \textit{radial basis function (RBF) collocation method of solving PDEs} with deep learning. RBF collocation method is widely used as a numerical solver for PDEs because of its inherent meshfree nature that provides geometric flexibility with scattered node layout \cite{bollig2012solution}. Another advantage of RBF-based methods is that they can be used to solve high-dimensional problems since they convert multiple space dimensions into virtually one-dimension \cite{fasshauer2007meshfree, wu2004meshless}. We couple this RBF collocation method with deep learning to make it practicable to the systems where the governing PDEs are unknown. We propose to learn the application of the differential operators on a chosen RBF using deep neural networks, which makes the method independent of initial and boundary conditions while also utilizing the meshfree, geometrically flexible, and spatial dimension-independent nature of the RBF collocation method. By choosing infinitely-smooth RBFs (e.g. Gaussian, multiquadric, etc.) with trainable shape parameters, our method can be employed in scenarios where data sites are sparsely distributed. 

We first show how the RBF collocation method for time-dependent linear PDEs can be integrated with deep neural networks to incorporate unknown linear PDEs. Next, we extend the architecture to learn dynamics driven by unknown nonlinear PDEs of the form (\ref{eqn:problem}). Finally, we show how the proposed method can be used to learn a prediction model for a coupled system involving multiple nonlinear PDEs. We evaluate our method in the forecasting task for the two-dimensional wave equation and the Burgers-Fisher equation and demonstrate the capability of transferring the learned model in different geometries with different boundary conditions. Furthermore, we show the applicability of our method for high-dimensional systems using the example of ten-dimensional heat equation. 

Our main contributions are the following:
\begin{itemize}
    \item We develop a method for learning prediction models for spatiotemporal physical processes driven by unknown PDEs from sparsely-observed data. 
    \item Our method is suitable for high-dimensional systems as well as independent of initial and boundary conditions. 
    \item The proposed method allows different locations of measurement sites, even different numbers of measurement sites, across multiple training and test sample sequences.
    \item The proposed method is geometry-independent. A model trained in one geometry can be transferred to different geometries with different boundary and initial conditions provided that the governing PDE remains the same.
\end{itemize}

%% file: backgroud.tex
\section{Related work and Preliminaries}

\subsection{Related work} 
Deep learning methods for learning or solving PDE-driven systems can be broadly categorized into two types. The first kind of method directly approximate the PDE solution with deep neural networks (DNNs) as a function of space and time \cite{raissi2019physics, sirignano2018dgm, raissi2018deep, li2019d3m}.
These methods are meshfree, geometrically flexible, and suitable for high dimensional PDEs. However, since these methods directly approximate the PDE solution as a function of space and time, the learned model is dependent on the specific initial and boundary condition used in training. Furthermore, these methods require either a large number of data sites or knowledge of the PDE for generalization. 

The second type of method learn the differential operators using convolutional neural networks (CNNs) \cite{long2018pde, de2019deep, long2019pde, bar2019learning}. These methods are independent of initial and boundary conditions, but they are not suitable for high-dimensional PDEs and can only be used when data is available in a dense regular grid. Furthermore, both types of methods are often focused on solving known PDEs and incorporate the PDEs either in the model structure \cite{long2018hybridnet, de2019deep, saha2020phicnet, sirignano2020dpm} or in the loss function \cite{raissi2019physics, sirignano2018dgm, geneva2020modeling} limiting their applicability in scenarios where the governing dynamics is unknown and needs to be learned from observed data.

One promising approach for solving or learning PDEs using irregular scattered data is to utilize graph neural networks (GNNs) \cite{scarselli2008graph}. Seo and Liu \cite{seo2019differentiable} proposed differentiable physics-informed graph networks (DPGN) which incorporates differentiable physics equations in GNN to model complex dynamical systems. However, difference operators on a graph can extract spatial variations only when the nodes are close to each other and contribute to large numerical errors when data sites are sparsely distributed \cite{seo2019physics}.

\subsection{Radial Basis Functions (RBF\lowercase{s}) Interpolation}
In standard scattered data interpolation problem, we are generally given a set of $N$ distinct data points $X = \{\mathbf{x}_1, \cdots, \mathbf{x}_N\} \subset \Omega \subset \mathbb{R}^d$ and corresponding real values (measurements) $y_i, i = 1, \cdots, N$. The goal is to find a (continuous) function $f: \mathbb{R}^d \rightarrow \mathbb{R}$ that satisfies the interpolation conditions
\begin{equation}
    f(\mathbf{x}_i) = y_i, \quad i = 1, \cdots, N
    \label{eqn:intrp_conds}
\end{equation}
A common approach to solve this scattered data interpolation problem is to consider the function $f(\mathbf{x})$ as a linear combination of basis functions of a certain class. Often these basis functions are formed using \textit{radial functions}. Value of a radial function $\phi: \mathbb{R}^d \rightarrow \mathbb{R}$ at each point depends only on some arbitrary distance between that point and origin such that $\phi(\mathbf{x}) = \phi(\mathbb{\|\mathbf{x}\|})$ where $\|\cdot\|$ denotes some norm in $\mathbf{R}^d$, usually the Euclidean norm. Effectively, the function becomes a univariate function $\phi: \mathbb{R}^+ \rightarrow \mathbb{R}$.

In radial basis functions interpolation, the basis functions are formed using the given datapoints $\mathbf{x}_i, i = 1, \cdots, N$ as centers or origins. In terms of these basis functions $\phi(\|\cdot\|)$, the interpolant takes the following form:   
\begin{equation}
    f(\mathbf{x}) = \sum_{j=1}^N c_j \phi(\|\mathbf{x} - \mathbf{x}_j\|), \quad \mathbf{x} \in \Omega
    \label{eqn:interpolant}
\end{equation}
where $c_j, j = 1, \cdots, N$ are the unknown coefficients to be determined. Plugging the interpolation conditions (\ref{eqn:intrp_conds}) into (\ref{eqn:interpolant}) leads to a system of linear equations
\begin{equation}
    \mathbf{f} = \Phi \mathbf{c},
    \label{eqn:linear_eqns}
\end{equation}
where $\Phi$ is the \textit{interpolating matrix} whose entries are given by $\Phi_{ij} = \phi(\|\mathbf{x}_i - \mathbf{x}_j\|), \  i,j = 1, \cdots, N$, $\mathbf{c}$ is the coefficient vector $[c_1, \cdots c_N]^T$ and $\mathbf{f} = [f(\mathbf{x}_1), \cdots, f(\mathbf{x}_N)]^T$.

Solution to the system (\ref{eqn:linear_eqns}) exists and it is unique provided the interpolating matrix $\Phi$ is non-singular. For a large class of radial functions including (inverse) multiquadrics, Gaussian, the matrix $\Phi$ is non-singular if the datapoints are all distinct \cite{fasshauer2007meshfree}. Details regarding the invertibility of the interpolating matrix for various choices of the radial functions can be found in \cite{buhmann2003radial, fasshauer2007meshfree}. 

\subsection{Solving time-dependent PDE\lowercase{s} using RBF\lowercase{s}}
\label{subsec: background_2}

Radial basis functions are widely used in meshfree methods for solving PDE problems. Here we will describe how RBFs are used, particularly Kansa's unsymmetric collocation method \cite{kansa1990multiquadrics}, to solve time-dependent linear PDEs. We develop our proposed approach based on this RBF collocation method.  

Consider the following time-dependent linear PDE:
\begin{equation}
    \partial_t u = Lu, \quad \mathbf{x} \in \Omega \subset \mathbb{R}^d, \quad t > 0,
    \label{eqn:linear_pde}
\end{equation}
with initial and boundary conditions
\begin{equation}
    u(0, \mathbf{x}) = w(\mathbf{x}), 
    \quad \mathbf{x} \in \Omega, 
    \label{eqn:linear_pde_ic}
\end{equation}
\begin{equation}
    u(t, \mathbf{x}) = g(t, \mathbf{x}) \quad \mathbf{x} \in \partial \Omega, \quad t > 0.
    \label{eqn:linear_pde_bc}
\end{equation}
$L: \mathbb{R} \rightarrow \mathbb{R}$ is some linear (spatial) differential operator, $\partial \Omega$ denotes the boundary of $\Omega$.

\vspace{2mm}
In order to obtain an approximate solution of the above PDE, consider $X = \{\mathbf{x}_1, \cdots, \mathbf{x}_N\} \subset \Omega$ be a set of $N$ collocation nodes.
Furthermore, let us assume that out of these $N$ collocation nodes, the last $b$ nodes are boundary nodes, i.e., the subset $X_B = \{\mathbf{x}_{N-b+1}, \cdots, \mathbf{x}_N\} \subset \partial \Omega$.
Using some radial basis functions $\phi(\|\cdot\|)$ centered at the collocation nodes, an approximate solution to the PDE can be defined as follows:
\begin{equation}
    \hat u(t, \mathbf{x}) = \sum_{j=1}^N c_j (t) \phi(\|\mathbf{x} - \mathbf{x}_j\|), \quad \mathbf{x} \in \Omega, \quad t \geq 0
    \label{eqn:linear_pde_radial_approximation}
\end{equation}
where $c_j(t), j = 1, \cdots, N$ are the unknown time-dependent coefficients to be determined at each time step. Plugging (\ref{eqn:linear_pde_radial_approximation}) into (\ref{eqn:linear_eqns}), (\ref{eqn:linear_pde_ic}), and (\ref{eqn:linear_pde_bc}) we get the following ordinary differential equation (ODE) 
\begin{align}
    \sum_{j=1}^N \frac{dc_j(t)}{dt}  \phi(\|\mathbf{x} - \mathbf{x}_j\|) - c_j(t) L\phi(\|\mathbf{x} - \mathbf{x}_j\|) = 0, \nonumber \\ \quad \mathbf{x} \in \Omega, \quad t > 0, 
    \label{eqn:ode}
\end{align}
with initial and boundary conditions
\begin{equation}
    \sum_{j=1}^N c_j (0) \phi(\|\mathbf{x} - \mathbf{x}_j\|) = u(0, \mathbf{x}), \quad \mathbf{x} \in \Omega, 
    \label{eqn:ode_ic}
\end{equation}
\begin{equation}
    \sum_{j=1}^N c_j (t) \phi(\|\mathbf{x} - \mathbf{x}_j\|) = g(t, \mathbf{x}), \quad \mathbf{x} \in \partial \Omega, \quad t > 0.
    \label{eqn:ode_bc}
\end{equation}
In (\ref{eqn:ode}), the spatial differential operator $L$ is applied to the radial basis function. Approximating the time-derivatives $\frac{dc_j(t)}{dt}$ by first order finite difference, we get
\begin{align}
    & \sum_{j=1}^N  c_j(t + \Delta t) \phi(\|\mathbf{x} - \mathbf{x}_j\|) \nonumber \\ 
    & = \sum_{j=1}^N c_j(t) \Big( \phi(\|\mathbf{x} - \mathbf{x}_j\|) + \Delta t L\phi(\|\mathbf{x} - \mathbf{x}_j\|)\Big), \nonumber \\ & \hspace{5cm} \mathbf{x} \in \Omega, \quad t \geq 0,
    \label{eqn:ode_fd}
\end{align}
where $\Delta t > 0$ is the step-size in time. The coefficients $c_j(t)$ can be determined by applying (\ref{eqn:ode_bc}) and (\ref{eqn:ode_fd}) on the boundary collocation nodes of $X_B$ and the interior collocation nodes of $X \setminus X_B$. In that case, (\ref{eqn:ode_ic}), (\ref{eqn:ode_bc}), and (\ref{eqn:ode_fd}) can be written in the following compact form:
\begin{align}
    \Phi \mathbf{c}^{(t+\Delta t)} &= A\mathbf{c}^{(t)} + \mathbf{g}^{(t+\Delta t)}, \quad t \geq 0 \nonumber \\ \Phi \mathbf{c}^{(0)} &= \mathbf{u}^{(0)}
    \label{eqn:compact}
\end{align}
$\Phi$ is the RBF matrix whose entries are given by $\Phi_{ij} = \phi(\|\mathbf{x}_i - \mathbf{x}_j\|)$, $\  i,j = 1, \cdots, N$ and $\mathbf{c}^{(t)} = [c_1(t), \cdots c_N(t)]^T$ is the coefficient vector at time $t$. The elements of the matrix $A$ are given by
\begin{align}
    A_{ij} = \begin{cases}
                 \phi(\|\mathbf{x}_i - \mathbf{x}_j\|) + \Delta t L\phi(\|\mathbf{x} - \mathbf{x}_j\|)|_{\mathbf{x} = \mathbf{x}_i}, \\ \hspace{24mm} if \quad 1 \leq i \leq (N-b), \  1 \leq j \leq N \\
                 0, \hspace{3mm} otherwise\  (\text{i.e. } (N-b) < i \leq N, \  1 \leq j \leq N)
             \end{cases}
\end{align}
and, $\mathbf{g}^{(t)} = [0, \cdots, 0, g(t,\mathbf{x}_{N-b+1}), \cdots, g(t,\mathbf{x}_N)]$ is the boundary condition vector at time $t$. $\mathbf{u}^{(0)} = [u(0, \mathbf{x}_1), \cdots, u(0,\mathbf{x}_N)]^T$ is the vector of initial values at the collocation nodes.
The coefficient vectors $\mathbf{c}^{(t)}$ are obtained by iteratively solving (\ref{eqn:compact}) which are used to compute the numerical solution $\hat u(t, \mathbf{x})$ by (\ref{eqn:linear_pde_radial_approximation}).

%% file: approach.tex
\section{Proposed Approach}

\begin{figure*}[t]
    \centering
    \subfloat[]{\includegraphics[width=0.65\linewidth]{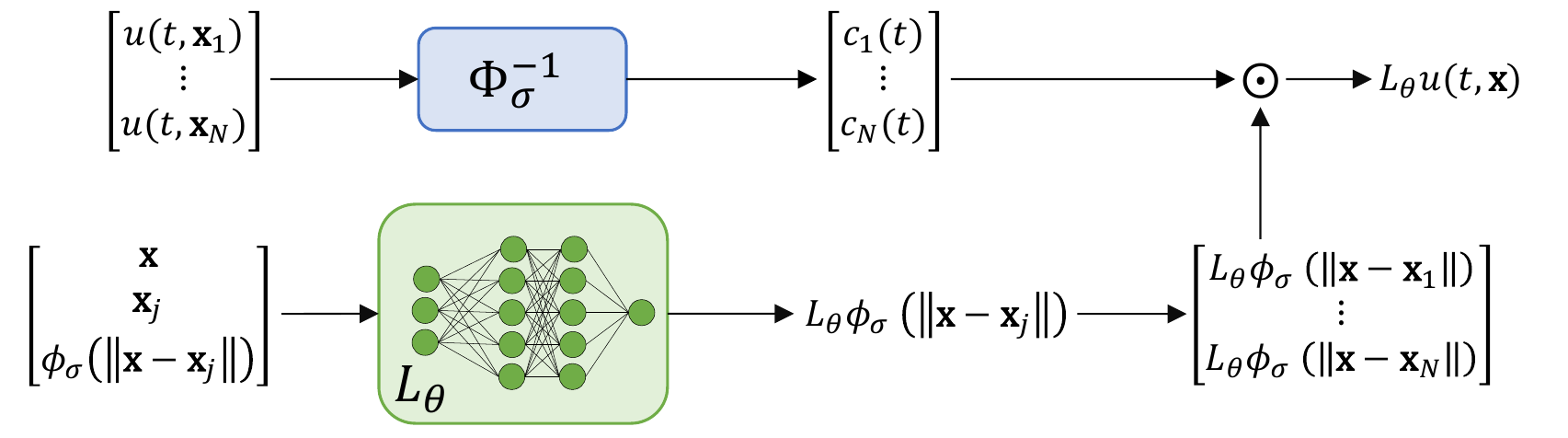}} \\
    \subfloat[]{\includegraphics[width=0.75\linewidth]{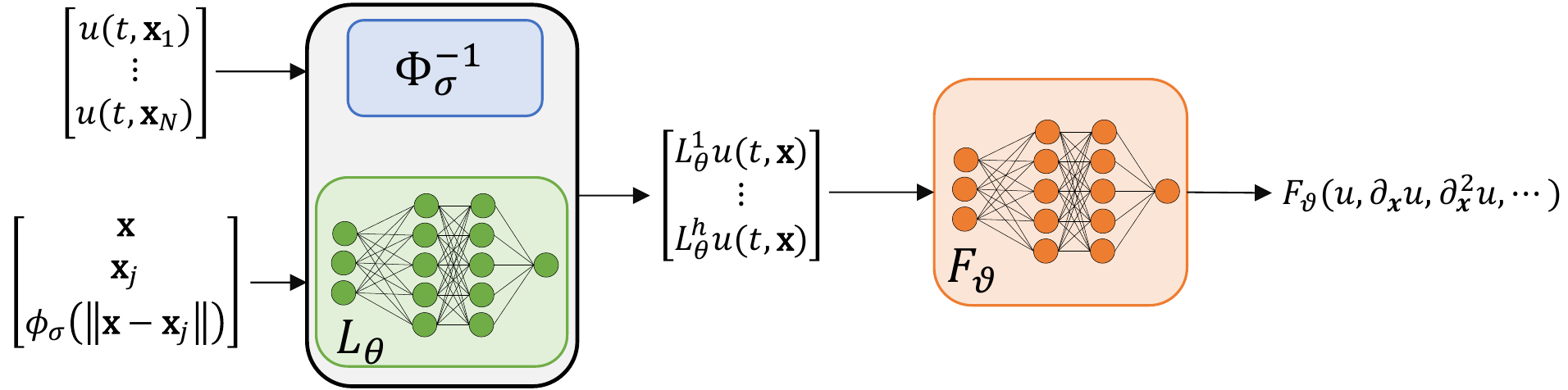}} \\
    \subfloat[]{\includegraphics[width=0.84\linewidth]{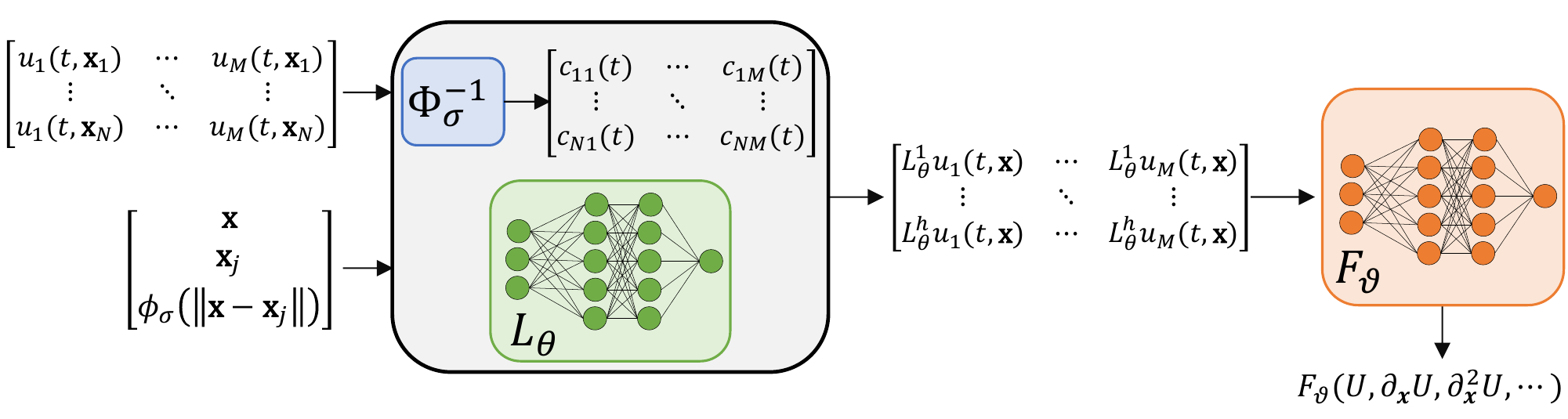}}
    \caption{(a): Proposed model for a system involving a linear PDE; $\Phi_\sigma$ is the RBF interpolation matrix with trainable shape parameter $\sigma$; the unknown linear differential operator $L$ is modeled using a feed-forward neural network $L_\theta$ with parameter vector $\theta$. (b): Proposed model for a system driven by a nonlinear PDE of the form (\ref{eqn:problem}); $L_{\theta, h}$ is same as $L_\theta$ except it has $h$ output neurons instead of just one; $F_{\vartheta}$ is the neural network representation (with parameter vector $\vartheta$)  of the unknown function $F$. (c): Proposed model for a system of $M$ coupled PDEs. $U(t, \mathbf{x})$ represents a vector containing the $M$ measurement variables $[u_1(t, \mathbf{x}), \cdots, u_M(t, \mathbf{x})]^T$.}
    \label{fig:models}
\end{figure*}

We will first describe our approach for unknown linear PDE-driven systems and then extend that to systems that involve nonlinear PDE of the form (\ref{eqn:problem}). Finally, we will show how the proposed model can be used to learn an unknown system of coupled nonlinear PDEs.

\subsection{Learning dynamics driven by unknown linear PDE}
Consider the linear time-dependent PDE of the general form shown in (\ref{eqn:linear_pde}), where the linear (spatial) differential operator $L$ is unknown. Learning the application of $L$ directly on $u$ requires fine-grid measurement for convolution operations. Rather, we propose to learn the application of $L$ on a radial basis function $\phi$ that allows scattered sparse measurements independent of any specific geometry. Specifically, we propose to learn a feed-forward neural network $L_{\theta}$, with parameter vector $\theta$, that approximates $L\phi(\|\mathbf{x} - \mathbf{x}_j\|)|_{\mathbf{x} = \mathbf{x}_i}$ given any two spatial locations $\mathbf{x}_i, \mathbf{x}_j$ and the corresponding value of the RBF $\phi(\|\mathbf{x}_i - \mathbf{x}_j\|)$. We train the neural network $L_{\theta}$ using scattered time-series measurement of $u$ by minimizing an objective function derived from (\ref{eqn:compact}). Let $X = \{\mathbf{x}_1, \cdots, \mathbf{x}_N\} \subset \Omega$ be the set of measurement sites and $\mathbf{u}^{(t)} = [u(t, \mathbf{x}_1), \cdots, u(t, \mathbf{x}_N)]^T$ be the measurement vector at time $t$. Since our goal is to learn a model for spatial differential operator $L$ and boundary values are not affected by $L$, we only consider internal measurement sites during training. For a radial basis function $\phi$, the coefficient vector at time $t$, $\mathbf{c}^{(t)}$, can be obtained by solving the linear system 
\begin{equation}
    \Phi \mathbf{c}^{(t)} = \mathbf{u}^{(t)}
    \label{eqn:linear_at_t}
\end{equation}
$\Phi$ is the RBF matrix whose entries are given by $\Phi_{ij} = \phi(\|\mathbf{x}_i - \mathbf{x}_j\|)$, $\  i,j = 1, \cdots, N$. Replacing $L$ with $L_{\theta}$, ignoring the boundary nodes, and plugging (\ref{eqn:linear_at_t}) into (\ref{eqn:compact}) we get
\begin{equation}
    \mathbf{u}^{(t+\Delta t)} =  \mathbf{u}^{(t)} + \Delta t D_{\theta}\Phi^{-1}_\sigma \mathbf{u}^{(t)} = (I + \Delta t D_{\theta}\Phi^{-1}_\sigma) \mathbf{u}^{(t)}
    \label{eqn:unknown_linear_pde_update}
\end{equation}
where $I$ is the identity matrix of order $N$, $D_\theta$ denotes the approximated spatial derivative matrix, i.e., $[D_\theta]_{ij} = L_{\theta}\phi(\|\mathbf{x} - \mathbf{x}_j\|)|_{\mathbf{x} = \mathbf{x}_i}$, $\  i,j = 1, \cdots, N$ and, $\sigma$ is the trainable shape parameter of the RBF. So, for a time-series of measurements $\mathbf{u}^{(t)} \in \mathbb{R}^N, t=t_0, t_0 + \Delta t,  \cdots, t_0 + K \Delta t$, the neural network $L_{\theta}$ is trained with the following objective:
\begin{align}
    \min_{\theta, \sigma} \frac{1}{K} \sum_{k=0}^{K-1} \big\|\mathbf{u}^{(t_0 + (k+1)\Delta t)} &- \mathbf{u}^{(t_0 + k\Delta t)} \nonumber \\ &- \Delta t D_\theta \Phi^{-1}_\sigma\mathbf{u}^{(t_0 + k\Delta t)}\big\|^2
\end{align}

Note, the term $D_\theta \Phi^{-1}\mathbf{u}^{(t)}$ in (\ref{eqn:unknown_linear_pde_update}) basically multiplies the approximated spatial derivative matrix $D_\theta$ with coefficient vector $\mathbf{c}^{(t)}$ to obtain the radial approximation of $L\mathbf{u}^{(t)}$. Once the model $L_\theta$ is trained, it can be used to get the value of $L_\theta u(t, \mathbf{x})$ (Figure \ref{fig:models}a) at any generic location $\mathbf{x} \in \Omega$ and to finally predict $u(t, \mathbf{x})$. During the prediction phase, the boundary condition is applied similarly as in (\ref{eqn:compact}).
Considering the last $b$ nodes in $X$ as boundary nodes and incorporating the corresponding boundary condition of (\ref{eqn:ode_bc}) in (\ref{eqn:unknown_linear_pde_update}), we get 
\begin{equation}
    \mathbf{u}^{(t+\Delta t)} = H \mathbf{u}^{(t)} + \mathbf{g}^{(t+\Delta t)},
    \label{eqn:prediction_eqn} 
\end{equation}
where the rows of matrix $H$ are given by
\begin{align}
    H_i = \begin{cases}
              [I + \Delta t D_\theta \Phi_\sigma^{-1}]_i & if \quad 1 \leq i \leq (N-b) \\
              \mathbf{0} & otherwise
          \end{cases}
\end{align}

Once the model is trained, during prediction, the matrix $H$ is solely determined from the configuration of measurement sites. Under a fixed Dirichlet boundary condition $g(t, \mathbf{x}) = g(\mathbf{x})$, the iterative scheme of (\ref{eqn:prediction_eqn}) is a linear discrete-time dynamical system and its convergence to an equilibrium point can be analyzed from the spectral properties of $H$. 
\newline
\textit{The linear discrete-time dynamical system $\mathbf{u}^{(t+\Delta t)} = H \mathbf{u}^{(t)} + \mathbf{g}$ converges to a stable equilibrium point from any arbitrary initialization if and only if $|\lambda_H| < 1$, where $\lambda_H$ are the eigenvalues of $H$.}

\subsection{Learning dynamics driven by unknown nonlinear PDE}
In order to learn an unknown nonlinear PDE of the form (\ref{eqn:problem}), we use the proposed model for linear PDEs as the base module. The input arguments of the nonlinear function $F$ in (\ref{eqn:problem}) are basically various unknown linear spatial derivatives of the measurement variable $u$. Therefore, we first use the linear model to generate a spatial derivative feature vector and then pass that through another feed-forward neural network $F_\vartheta$,  with parameter vector $\vartheta$, that approximates the function $F$. The neural network $L_\theta$ is modified to generate a vector of length $h$ (instead of a scalar as in the case of linear PDEs), whose entries represent spatial derivative features of the RBF $\phi$. We denote this spatial derivative feature extractor neural network as $L_{\theta, h}$. The output of $L_{\theta, h}$ is multiplied with the coefficient vector $\mathbf{c}^{(t)}$ to get the spatial derivative features of $u$, which is used by $F_\vartheta$ to obtain the final output. Figure \ref{fig:models}b shows the overall model for nonlinear PDEs. In this case, for a time-series of measurements (at internal nodes) $\mathbf{u}^{(t)} \in \mathbb{R}^N, t=t_0, t_0 + \Delta t,  \cdots, t_0 + K \Delta t$, the neural networks $L_{\theta, h}$ and $F_\vartheta$ are trained jointly with the following objective:
\begin{align}
    \min_{\theta, \vartheta, \sigma} \frac{1}{K} \sum_{k=0}^{K-1} \big\|&\mathbf{u}^{(t_0 + (k+1)\Delta t)} - \mathbf{u}^{(t_0 + k\Delta t)} \nonumber \\ &- \Delta t  F_\vartheta \big( D_{\theta, h} \Phi^{-1}_\sigma\mathbf{u}^{(t_0 + k\Delta t)}\big)\big\|^2,
    \label{eqn:nonlinear_objective_1}
\end{align}
where $D_{\theta, h} \in \mathbb{R}^{h \times N \times N}$ is a 3D tensor containing the $h$ spatial derivative matrices of the RBF. Note, the objective (\ref{eqn:nonlinear_objective_1}) considers nonlinear PDE of form (\ref{eqn:problem}) where the temporal order is $p=1$. In general, for $p \geq 1$, we use the finite difference approximation of $\frac{\partial^p}{\partial t^p}$ and the objective is modified as 
\begin{align}
    \min_{\theta, \vartheta, \sigma} \frac{1}{K} \sum_{k=0}^{K-1} \big\|& \mathbf{u}^{(t_0 + (k+1)\Delta t)} - \Delta t  F_\vartheta \big( D_{\theta, h} \Phi^{-1}_\sigma\mathbf{u}^{(t_0 + k\Delta t)}\big)  \nonumber \\ & - \textstyle \sum_{q=1}^{p} \binom{p}{q} (-1)^{q+1} \mathbf{u}^{(t_0 + (k-q+1)\Delta t)}\big\|^2
    \label{eqn:nonlinear_objective_p}
\end{align}

The training objective of (\ref{eqn:nonlinear_objective_1}) or (\ref{eqn:nonlinear_objective_p}) uses measurements from all sites to predict the values of $F$ at those sites but does not incorporate any constraints to generalize $F_\vartheta$ at locations where measurement is not available. Accuracy of RBF-based methods at any generic location $\mathbf{x} \in \Omega$ strongly depends on the shape parameter $\sigma$ of the RBF. Generally, the shape parameter is chosen using \textit{leave-one-out cross-validation} method \cite{rippa1999algorithm, fasshauer2007choosing}. Note, as the spatial differential operators are functions of $\sigma$, the parameters of $L_{\theta,h}$ are dependent on $\sigma$. We adopt the leave-one-out strategy in our training algorithm. At each training step, we randomly leave one measurement site and the model uses the remaining $(N-1)$ sites for prediction. The left-out measurement site is used only for computing the loss function. The overall training procedure is summarized in \textit{Algorithm \ref{algo:training}}.

\begin{figure}[t]
\removelatexerror
\begin{algorithm2e}[H]
  \caption{Training Procedure}
  \label{algo:training}
  \algorithmfootnote{$\mathbf{u}^{(t)}_{-l} \in \mathbb{R}^{N-1}$ denotes the vector with $l^{th}$ element removed from $\mathbf{u}^{(t)}$}
  
  \KwIn{$Z$ number of training samples \{ \newline 
  \-\hspace{14mm} measurement sites $X$, \newline 
  \-\hspace{14mm} measurement sequence $\mathbf{u}^{(t)},$  \newline
  \-\hspace{14mm} $\  t=t_0, t_0 + \Delta t,  \cdots, t_0 + K \Delta t$
  \}$^Z$, \newline
  a radial function $\phi$}
  \BlankLine 
  Arbitrarily initialize parameters $\theta, \vartheta, \sigma$
  \BlankLine 
  \While{not converged}{
  Randomly select sample sequence \{$X, \mathbf{u}^{(t)}, t=t_0, t_0 + \Delta t,  \cdots, t_0 + K \Delta t$\}
  \BlankLine 
  Randomly choose $l$ from $\{1, \cdots, N\}$
  \BlankLine 
  Compute RBF matrix $\Phi_\sigma \in \mathbb{R}^{N \times (N-1)}$ such that $[\Phi_\sigma]_{ij} =  \phi_\sigma(\|\mathbf{x}_i - \mathbf{x}_j\|), \ 1 \leq i, j \leq N, \ j \neq l $
  \BlankLine 
  Forward pass $L_{\theta, h}$ to get $D_{\theta, h} \in \mathbb{R}^{h \times N \times (N-1)}$ such that $[D_{\theta,h}]_{rij} = L_{\theta, h}^r \phi(\|\mathbf{x} - \mathbf{x}_j\|)|_{\mathbf{x} = \mathbf{x}_i}, \newline
  \hspace*{\fill} 1 \leq r \leq h, \ 1 \leq i, j \leq N, \ j \neq l $
  \BlankLine 
  Forward pass $F_\vartheta$ to get $F_\vartheta \big( D_{\theta, h} \Phi^{-1}_\sigma \mathbf{u}^{(t)}_{-l} \big) \in \mathbb{R}^N, \  t=t_0, t_0 + \Delta t,  \cdots, t_0 + K \Delta t$
  \BlankLine 
  Update $\theta, \vartheta, \sigma$ in the gradient descent direction of the objective (\ref{eqn:nonlinear_objective_p})
  }
  \BlankLine
  \KwOut{Learned parameters $\theta, \vartheta, \sigma$}
\end{algorithm2e}
\vspace{-10pt}
\end{figure}

Once the models $L_{\theta, h}$ and $F_\vartheta$ are trained, they can be used to get the approximated value of $F$ (Figure \ref{fig:models}b) at any generic location $\mathbf{x} \in \Omega$ and to finally predict $u(t, \mathbf{x})$.

\subsection{Learning unknown system of coupled nonlinear PDE\lowercase{s}}
Consider a coupled system of $M$ nonlinear PDEs associated with $M$ measurement variables $u_1, \cdots, u_M$. In order to learn such a system, we require radial approximation for each of these $M$ measurement variables. Therefore, given $N$ measurement sites, we consider an RBF coefficient matrix $C^{(t)} \in \mathbb{R}^{N \times M}$ (at time $t$) where each column corresponds to the coefficient vector of one measurement variable. Since the neural network $L_{\theta, h}$ operates only on spatial locations, its parameters are shared across all measurement variables. The output of $L_{\theta, h}$, i.e., the spatial derivative features of the RBF $\phi$, is multiplied with coefficient matrix $C^{(t)}$ to obtain the spatial derivative features of length $h$ for each of the measurement variables. These spatial derivative features are concatenated and finally passed through the neural network $F_\vartheta$ to get the final vector-valued ($\in \mathbb{R}^M$) output. Figure \ref{fig:models}c shows the overall model for a coupled system of nonlinear PDEs.

%% file: experiments.tex
\section{Simulation Results}

\begin{figure}[b]
    \centering
    \includegraphics[width=1\linewidth]{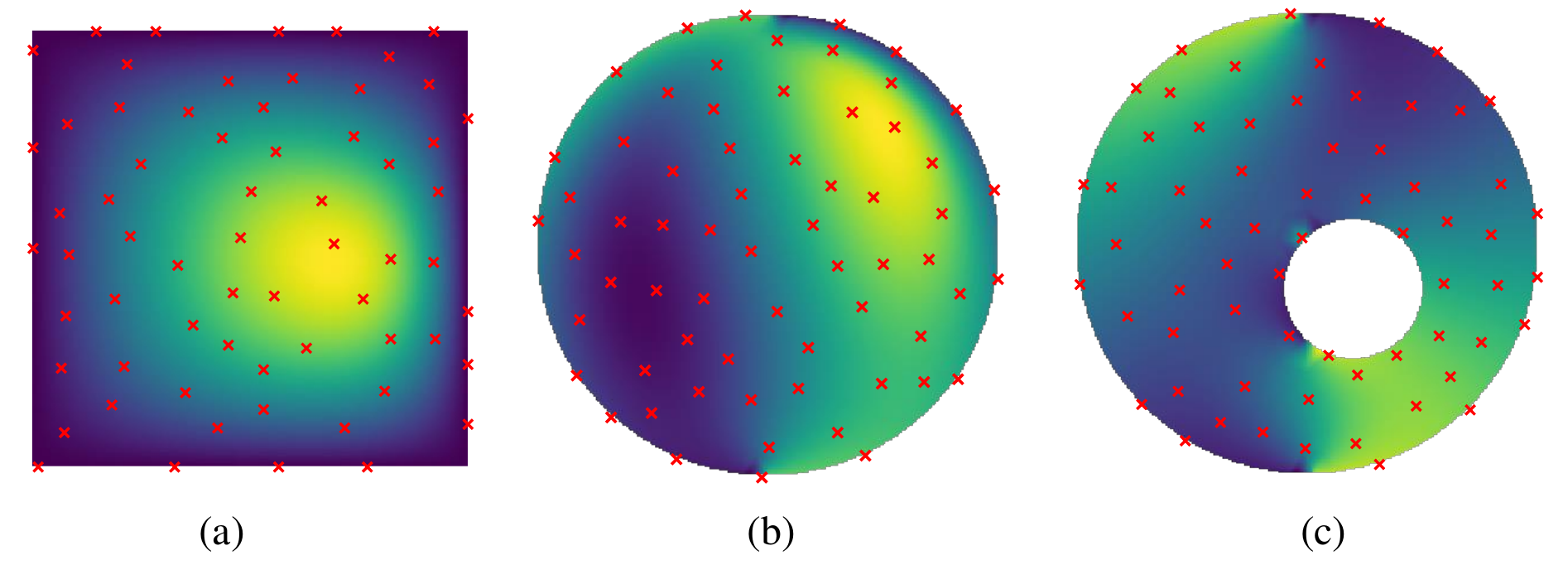}
    \caption{(a): Training geometry, (b, c): two additional geometries used for testing. Red markers show example measurement sites chosen randomly, maintaining a minimum pairwise distance.}
    \label{fig:geometries}
\end{figure}

\begin{figure}[t]
    \centering
    \includegraphics[width=0.7\linewidth]{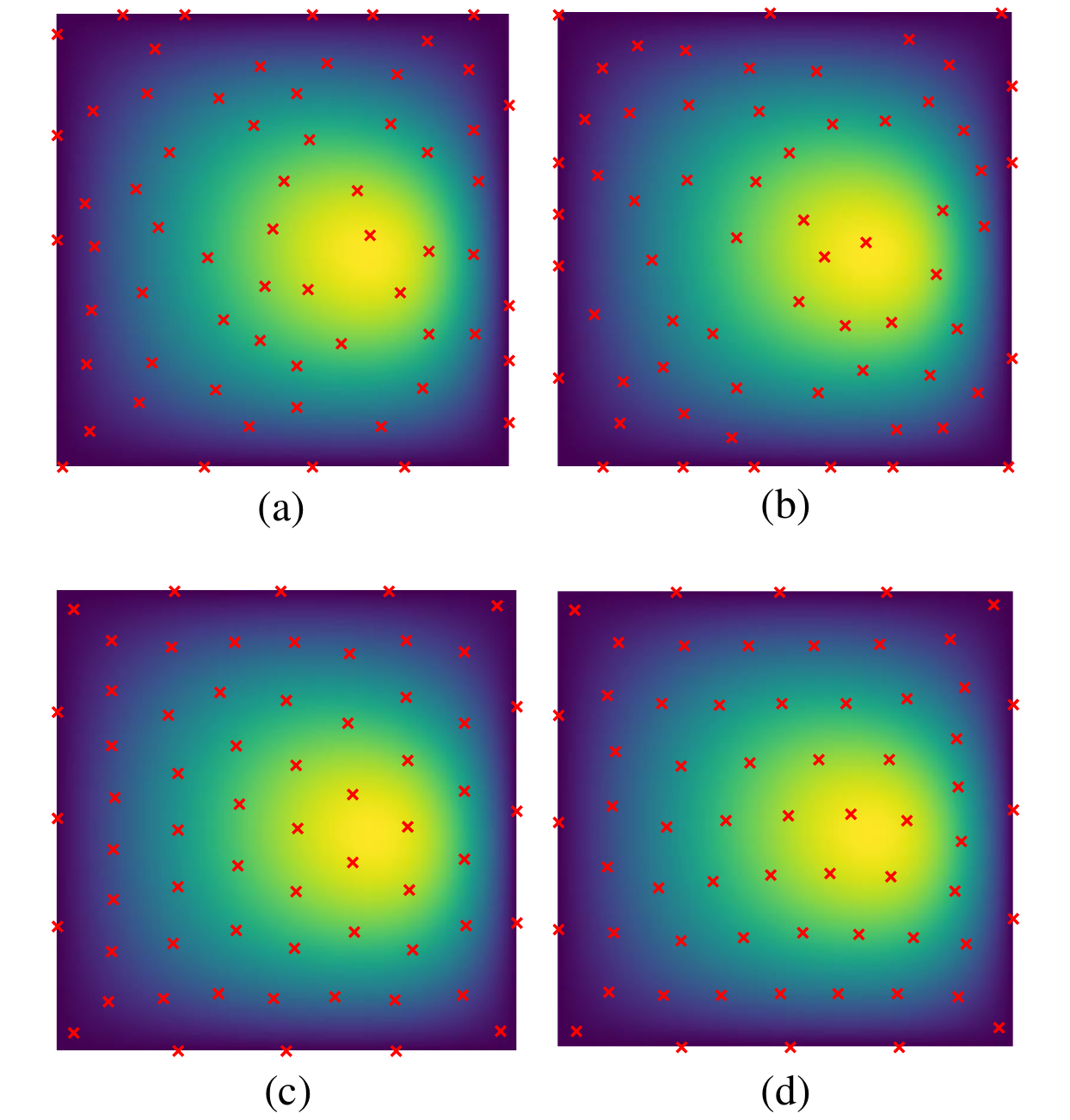}
    \caption{(a,b): Two examples of measurement site configuration on the square geometry obtained using random selection, maintaining a minimum distance among the measurement sites. (c,d): two examples of measurement site configuration on the square geometry obtained using K-means clustering.}
    \label{fig:sites}
\end{figure}

We evaluate the proposed method for the two-dimensional wave equation and the Burgers-Fisher equation. The first one is a linear PDE, whereas the latter is a coupled system of two nonlinear PDEs. Finally, we demonstrate the applicability of our method for high-dimensional system using the example of ten-dimensional heat equation. 

\subsection{Training and testing settings}
For the two-dimensional examples, we train the models with data from discrete measurement sites on a square geometry $[-1, 1] \times [-1, 1]$. Dataset is generated using \textit{finite element method} with fine-mesh consisting of roughly $1500$ nodes (depends on the geometry). 
For testing, we consider four different settings:
\renewcommand{\labelenumi}{(\roman{enumi})}
\begin{enumerate}
    \item same square geometry (used for training) with the same number of measurement sites (though their locations are different)
    \item same square geometry with different number of measurement sites
    \item a circular geometry with different boundary condition
    \item an annular geometry with different boundary condition
\end{enumerate}
Different geometries with example measurement site configurations are shown in Figure \ref{fig:geometries}. We investigate two different methods for choosing the measurement sites. In one case, measurement sites are chosen using \textit{K-means} clustering. In the other case, measurement sites are chosen randomly such that any pair of measurement sites are at least a minimum distance apart. Examples of measurement site distribution obtained using the two methods on the square geometry are shown in Figure \ref{fig:sites}. The number of measurement sites for our predictions is chosen to be within $4\%-5\%$ (depending on the geometry) of the number of nodes used for the ground truth FEM solution.

\subsection{Model and optimization hyperparameters}
We use the Gaussian function $\phi_\sigma (\|\mathbf{x}\|) = e^{-\|\mathbf{x}\|^2 / 2 \sigma^2}$ as the radial function, where $\sigma$ is the shape parameter.
\newline
For $L_{\theta, h}$ and $F_\vartheta$, we use fully connected networks. 
The fully-connected network for $L_{\theta, h}$ consists of two hidden layers with $64$ and $32$ neurons, respectively, followed by ReLU activations. The numbers of input and output neurons are $(2d+1)$ and $h$, respectively, where
$d$ is the dimension of the domain ($d=2$ for the wave and Burgers-Fisher examples, and $d=10$ for the heat example). $h$ denotes the length of the spatial derivative feature vector; we use $h=16$ for all examples. 
The fully-connected network for $F_\vartheta$ comprises three hidden layers with $128$, $64$, and $32$ neurons, respectively, followed by ReLU activations. The numbers of input and output neurons are $Mh$ and $M$, respectively. $M$ is the number of PDEs present in the system; $M=1$ for the wave equation and heat equation, whereas $M=2$ for the Burgers-Fisher equation. 

We use $300$ sequences of length $200$ for training and validation; another $50$ sequences are used for testing. We train the models for $200$ epochs in mini-batches of $32$ using Adam optimizer \cite{kingmaB14} with an initial learning rate of $0.001$. 
During prediction, at each timestep, the linear system $\Phi \mathbf{c}^{(t)} = \mathbf{u}^{(t)}$ is solved using $\ell_2$-regularized least square (with a regularization parameter $= 10^{-4}$) instead of direct inversion to avoid blowing up coefficients due to noisy estimates. Models are implemented, trained, and tested in PyTorch \cite{paszke2019pytorch}.

\subsection{Evaluation process and metrics}
We evaluate our method on the long-term prediction task. Only a few initial sparse observations are used for the very first prediction. Thereafter, the sparse prediction at each step is iteratively used by the model for long-term prediction, no intermediate observation is used (Figure \ref{fig:intro}).

We use Signal-to-Noise Ratio (SNR) as a metric to evaluate the prediction quality. Depending on the scenario, we may seek prediction only on the measurement sites $X$ or on the entire domain $\Omega$. Therefore, we define two SNR metrics as follows.
\begin{align}
    \text{SNR}_X(u, \hat u, t) &= 10 \log_{10} \frac{\sum_{\mathbf{x} \in X} u^2(t, \mathbf{x})}{\sum_{\mathbf{x} \in X} (u(t, \mathbf{x}) - \hat u(t, \mathbf{x}))^2} \\
    \text{SNR}_{\Omega}(u, \hat u, t) &= 10 \log_{10} \frac{\sum_{\mathbf{x} \in \Omega} u^2(t, \mathbf{x})}{\sum_{\mathbf{x} \in \Omega} (u(t, \mathbf{x}) - \hat u(t, \mathbf{x}))^2}
\end{align}
$\hat{u}(t, \mathbf{x})$ denotes the predicted value of $u$ at time $t$, at location $\mathbf{x}$. For the wave equation and the Burgers-Fisher equation examples, an analytical solution is not feasible. Therefore, we treat the fine-mesh FEM solution as the ground truth ($u$) to evaluate the prediction accuracy.  

\begin{figure}[b]
    \centering
    \includegraphics[width=0.7\linewidth]{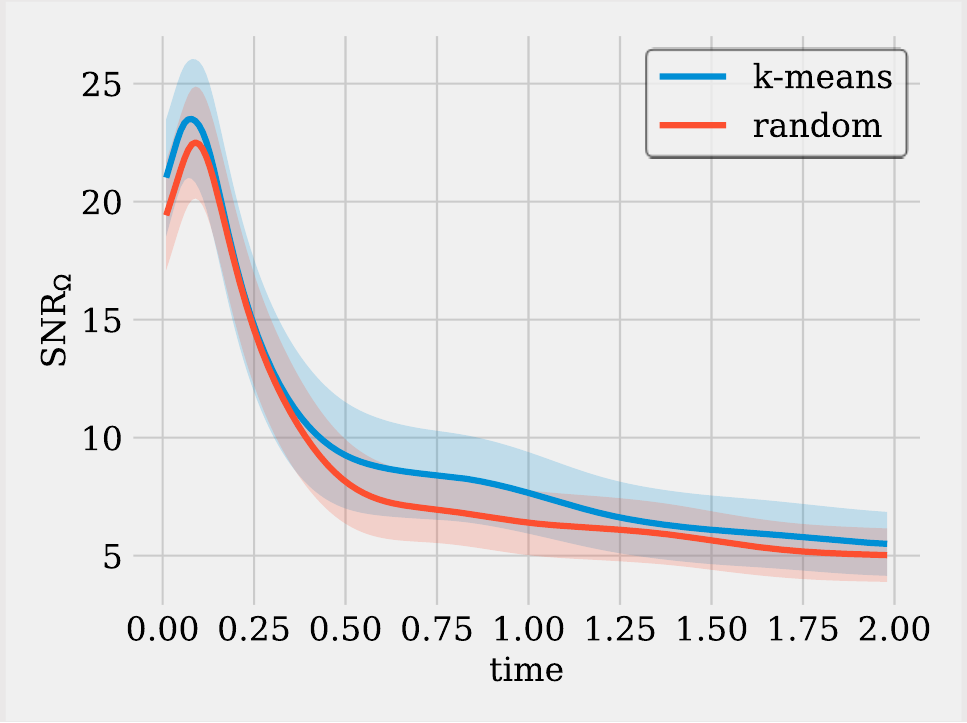}
    \caption{SNR comparison for the wave equation example on settings (i) when the measurement sites are chosen using K-means clustering versus when the measurement sites are chosen randomly, maintaining a minimum pairwise distance.}
    \label{fig:kmeans-vs-random}
\end{figure}

\begin{figure*}[!ht]
    \centering
    \includegraphics[width=0.95\linewidth]{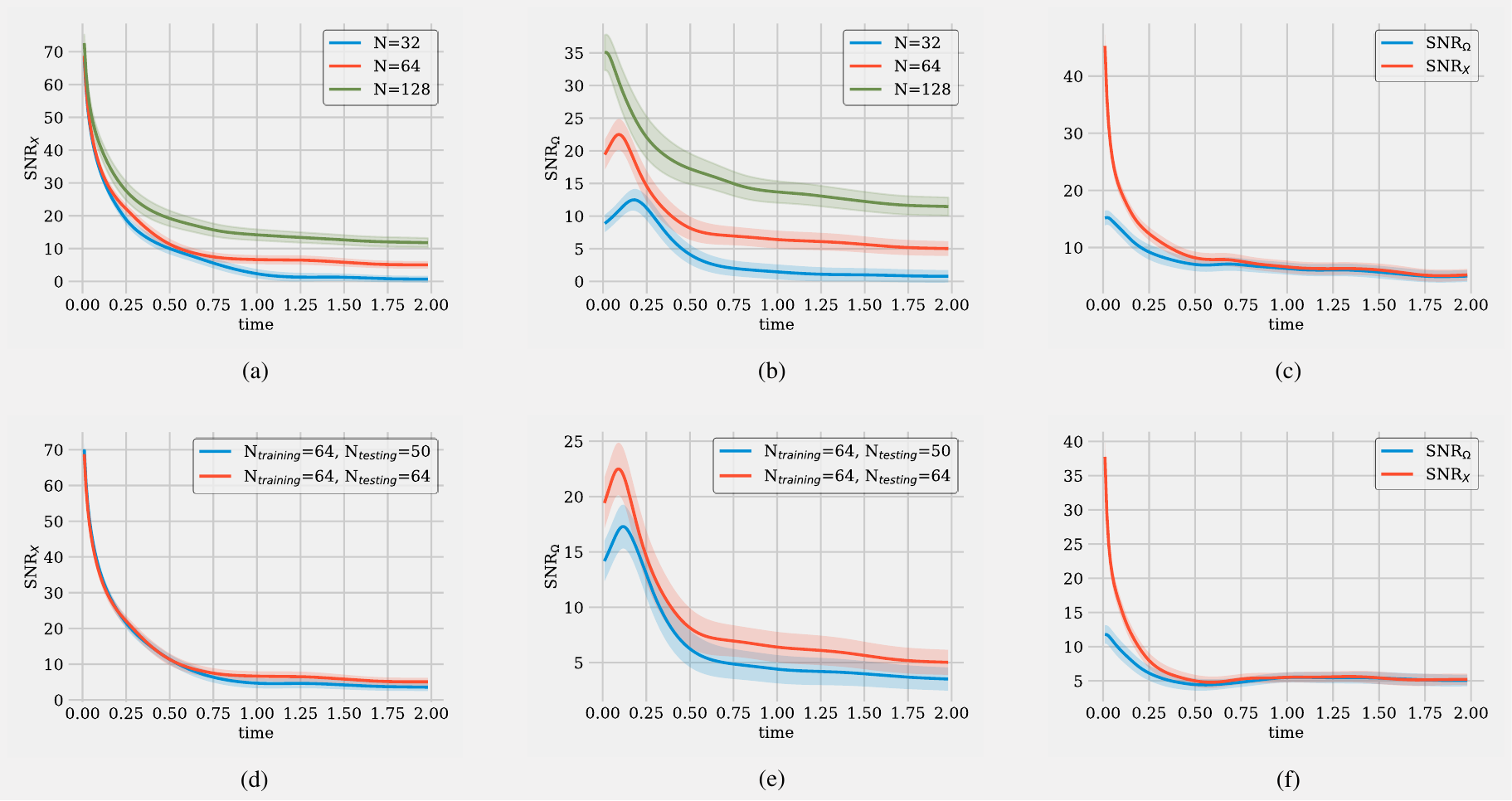}
    \caption{Quantitative analysis for the wave equation example. (a, b): SNR comparison in settings (i) with different number of measurement sites; same number of measurement sites is used in both training and testing. (d, e): SNR comparison in settings (ii). (c): SNR comparison in settings (iii). (f): SNR comparison in settings (iv). SNR$_X$ and SNR$_{\Omega}$ respectively denote SNR only at the measurement sites $X$ and on the entire domain $\Omega$.}
    \label{fig:wave-snr}
    \vspace{8mm}
    \includegraphics[width=0.95\linewidth]{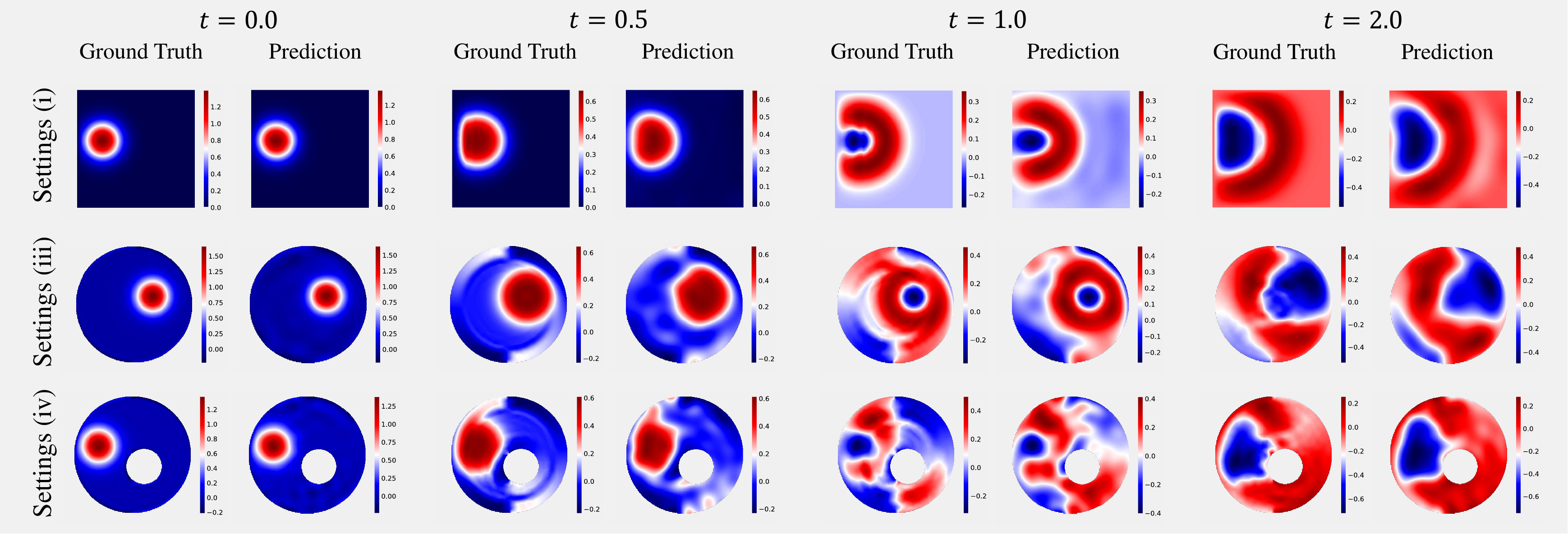}
    \caption{Visualization of the true and predicted states (on the entire domain $\Omega$) of the wave equation over time in different settings.}
    \label{fig:wave-vis}
\end{figure*}

\begin{figure*}[!ht]
    \centering
    \includegraphics[width=0.95\linewidth]{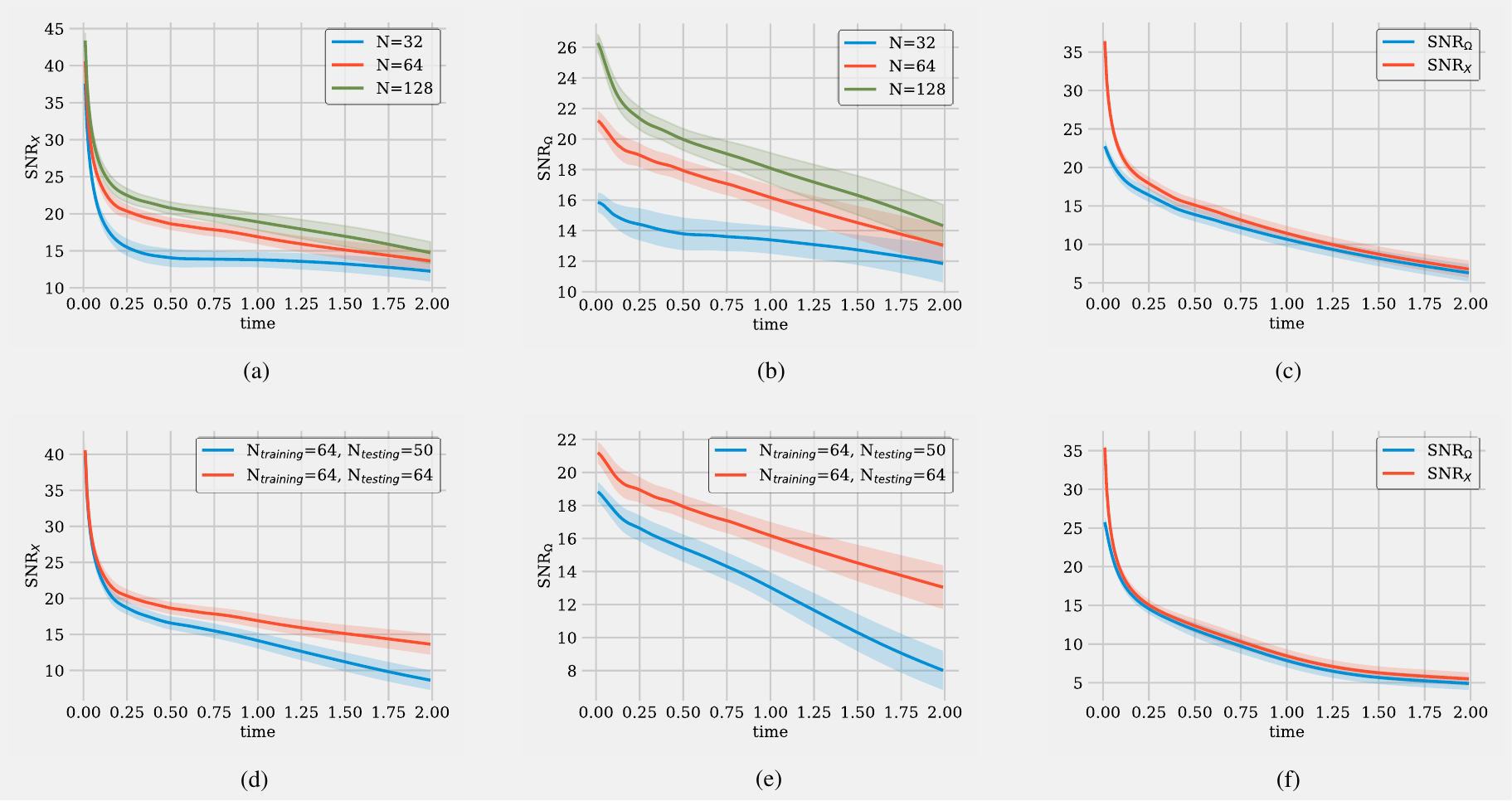}
    \caption{Quantitative analysis for the Burgers-Fisher equation example. (a, b): SNR comparison in settings (i) with different number of measurement sites; same number of measurement sites is used in both training and testing. (d, e): SNR comparison in settings (ii). (c): SNR comparison in settings (iii). (f): SNR comparison in settings (iv).}
    \label{fig:bf-snr}
    \vspace{8mm}
    \includegraphics[width=0.95\linewidth]{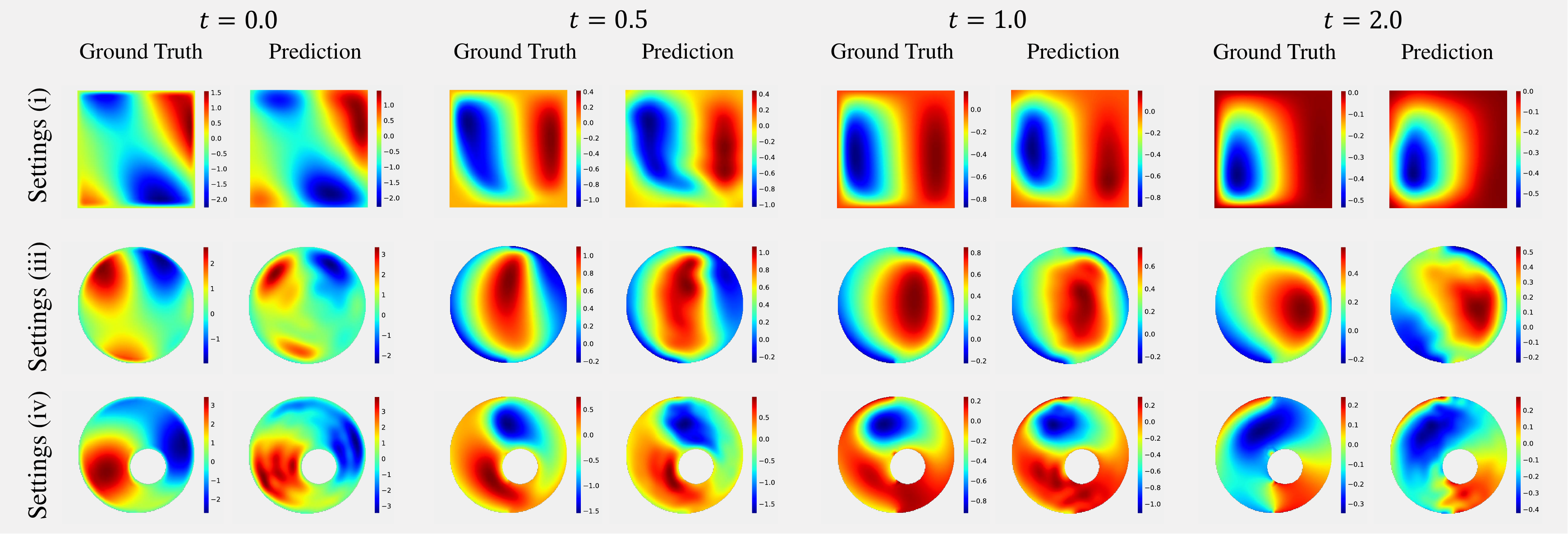}
    \caption{Visualization of the true and predicted states (on the entire domain $\Omega$) of one measurement variable of 2D Burgers-Fisher equation over time in different settings.}
    \label{fig:bf-vis}
\end{figure*}

\subsection{Wave equation}
We consider the two-dimensional wave equation given by 
\begin{equation}
    \partial_t^2 u  = \mathsf{v}^2 \nabla^2 u, \quad \mathbf{x} \in \Omega \subseteq [-1, 1] \times [-1, 1], \quad t \in [0, 2], 
\end{equation}
where $\mathsf{v}$ is the (constant) wave propagation speed and $\nabla^2$ denotes the differential operator Laplacian. 
We use $\mathsf{v} = 0.1$, and the following initial condition,
\begin{equation}
    u(0, \mathbf{x}) = a e^{-\varepsilon \|\mathbf{x} - \mathbf{z}\|^2},
    \label{eqn:gaussian_ic}
\end{equation}
where $a \sim \mathcal{U}(1, 2)$, $\varepsilon \sim \mathcal{U}(10, 100)$, and $\mathbf{z} \sim \mathcal{U}(\Omega)$ are chosen randomly for each sequence. $\mathcal{U}$ denotes the uniform distribution. 
For training dataset and test settings (i) and (ii), we use the Dirichlet boundary condition $u(t, \mathbf{x}) = 0, \mathbf{x} \in \partial \Omega, t > 0$. For test settings (iii) and (iv), the following boundary condition is used:
\begin{equation}
    u(t, \mathbf{x}) = 0.2 \  \text{sin}\Big(\text{tan}^{-1} \frac{x_2}{x_1}\Big), \quad \mathbf{x} = [x_1, x_2]^T \in \partial \Omega, \quad t > 0
    \label{eqn:diff_bc}
\end{equation}

Before going into the detailed results for this wave equation example, we show the effect of the measurement site selection process, K-means clustering versus random selection, on prediction accuracy. Figure \ref{fig:kmeans-vs-random} shows that the two methods lead to comparable SNR in settings (i). We observe a similar trend in other settings as well and for the rest of the quantitative results, we only show the figures that are obtained using the random selection method. 

Figure \ref{fig:wave-snr} shows the quantitative comparison of the proposed method in different settings in the task of forecasting. Increasing the number of measurement sites improves SNR (Figures \ref{fig:wave-snr}(a, b)). Comparing Figures \ref{fig:wave-snr}a and \ref{fig:wave-snr}b, it can be seen that the difference between prediction accuracy at the measurement sites only (SNR$_X$) and prediction accuracy on the entire domain (SNR$_{\Omega}$) decreases with time. Initially, the neural networks compensate for the spatial approximation error of RBFs to some extent and provide better accuracy at the measurement sites since they are trained only at those sites. However, as the prediction quality of the neural networks degrades over time, both the SNRs become very similar. Figures \ref{fig:wave-snr}(d, e) show that the prediction quality is maintained even if we use a different number of measurement sites compared to the training scenario. However, a larger change in the number of measurement sites would require retuning the RBF shape parameter. Figures \ref{fig:wave-snr}(c, f) show that the learned model can be used in different settings (different geometries with different boundary conditions). The model is trained in just one geometry with one boundary condition leading to a drop in accuracy when tested in other settings. The generalizability of the learned model can be improved by training it in multiple settings. A qualitative comparison of the predicted maps in different settings along with
the ground truth is depicted in Figure \ref{fig:wave-vis}.

\subsection{Burgers-Fisher equation}
We consider the two-dimensional coupled Burgers-Fisher equations given by 
\begin{align}
    \partial_t u_1 + [u_1, u_2]^T   \bigcdot \nabla u_1 &= \nu \nabla^2 u_1 + \alpha u_1 (1 - u_1), \nonumber \\
    \partial_t u_2 + [u_1, u_2]^T \bigcdot \nabla u_2 &= \nu \nabla^2 u_2 + \alpha u_2 (1 - u_2), \nonumber \\
    \mathbf{x} \in \Omega \subseteq [-1, 1] & \times [-1, 1], \quad t \in [0, 2], 
\end{align}
where $\nu$ and $\alpha$ are constant non-negative parameters, and $\nabla$ denotes the gradient operator. We use $[\cdot]^T$ to denote a column vector, whereas $[\cdot]$ represents a closed interval. 
We use $\nu = 0.1, \alpha=1$, and the following initial condition, 
\begin{align}
    u_m(0, \mathbf{x}) = \sum_{|\omega|,|\beta| < 4} & \lambda_{\omega, \beta} \ \text{cos}([\omega, \beta]^T \bigcdot \mathbf{x}) \nonumber \\ &+ \gamma_{\omega, \beta} \ \text{sin}([\omega, \beta]^T \bigcdot \mathbf{x}),  \quad m = 1, 2, \qquad&
\end{align}
where $\lambda_{\omega, \beta}, \gamma_{\omega, \beta} \sim \mathcal{N}(0, 0.2)$ are chosen randomly for each sequence. $\mathcal{N}$ denotes the normal distribution, and $\bigcdot$ denotes dot product. For training dataset and test settings (i) and (ii), we use the Dirichlet boundary condition $u_m(t, \mathbf{x}) = 0, m = 1, 2, \  \mathbf{x} \in \partial \Omega, \  t > 0$. For test settings (iii) and (iv), we use the same boundary condition as (\ref{eqn:diff_bc}). Measurements in real environments are often noisy; we add Gaussian noise $\mathcal{N}(0, 0.01 \times \text{SD}(u))$ to the generated data, where $\text{SD}(u)$ is the standard deviation of the clean training data.

Figure \ref{fig:bf-snr} shows the quantitative comparison of the proposed method in different settings in the task of forecasting. In this case, improvement in prediction accuracy due to changes in the number of measurement sites is relatively less (Figure \ref{fig:bf-snr}(a,b)). Convergence of the system to a smooth state (as opposed to wave system) can be attributed to the saturation in RBF approximation with a relatively less number of measurement sites. A qualitative comparison of the predicted maps in different settings along with the ground truth is depicted in Figure \ref{fig:bf-vis}.

\subsection{Heat equation}
\begin{figure}[t]
    \centering
    \includegraphics[width=0.7\linewidth]{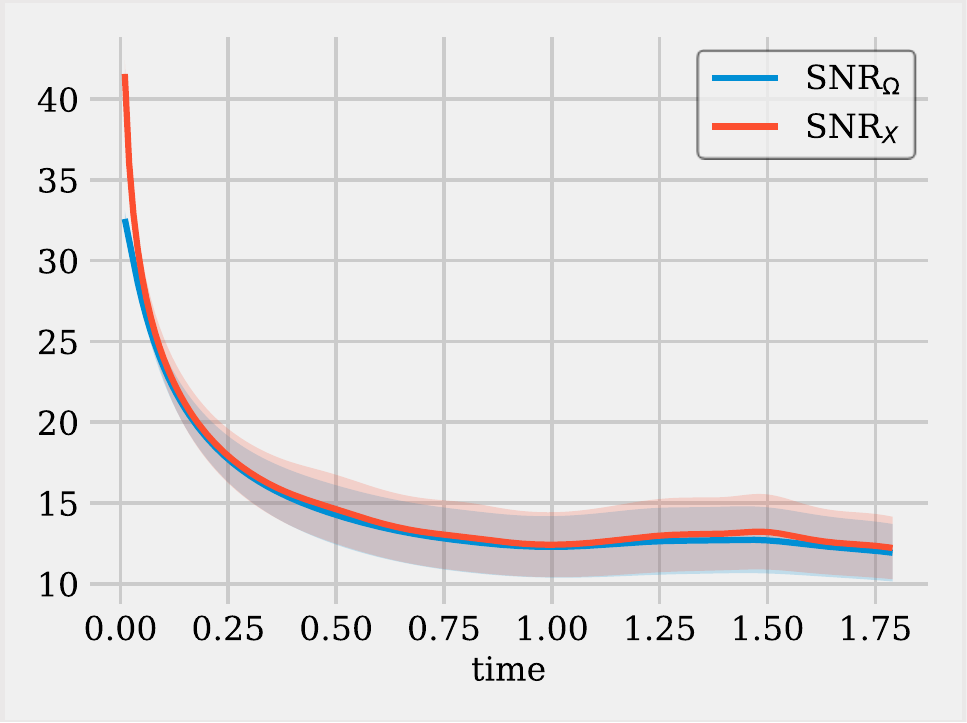}
    \caption{SNR of the predicted maps with respect to the analytical solution of the considered ten-dimensional heat equation.}
    \label{fig:heat-snr}
\end{figure}

Learning PDE-driven dynamics using deep learning in high-dimensional space has its own challenge. Obtaining data for training requires solving  high-dimensional PDEs, which itself is an open field of research \cite{han2020algorithms} and beyond the scope of this paper. Therefore, we choose an example where an analytical solution to the governing PDE is feasible. We consider $d=10$ dimensional heat equation  
\begin{equation}
    \partial_t u  = \kappa \nabla^2 u, \quad \mathbf{x} \in \Omega \subseteq [-0.2, 0.2]^d, \quad t \in [0, 2], 
\end{equation} with the Gaussian initial condition of (\ref{eqn:gaussian_ic}). $\kappa$ denotes the thermal diffusivity of the medium. Analytical solution of the considered PDE is given by
\begin{equation}
    u(t, \mathbf{x}) = \frac{ae^\frac{-\varepsilon \|\mathbf{x} - \mathbf{z}\|^2}{1 + 4 \kappa \varepsilon t}}{(1 + 4 \kappa \varepsilon t)^{d/2}}
    \label{eqn:heat_analytical}
\end{equation}
We obtain sequential data at randomly sampled sites using the analytical solution of (\ref{eqn:heat_analytical}) to train our model. Prediction from our model is quantitatively compared with respect to the analytical solution in Figure \ref{fig:heat-snr}.


%% file: conclusion.tex
\section{Conclusion and Future Work}
We developed a framework for learning prediction models for unknown PDE-driven physical processes from sparse observations by integrating the RBF collocation method for solving PDEs with deep learning. RBF framework enables meshfree computation and makes the method viable for learning in high-dimensional space. Learning the application of differential operators in a time-stepping fashion allows the learned model to be transferable to different settings of initial and boundary conditions. 

The current work focuses on presenting the fundamental framework of the approach in an application-agnostic way. The application of this fundamental method to a specific real example will require application-specific modifications. The current method assumes that the physical parameters (e.g. diffusion coefficient, wave speed) of the underlying system are uniform. However, in real scenarios, these physical parameters vary across space and time. Furthermore, real-world processes are often perturbed by various environmental forces. These factors should be incorporated in the current method before it is applied to learn real-world processes from sensor data. We would like to investigate this problem as future work.

In this work, we assume that direct observations of the state are available as data. However, in many cases, we can only obtain indirect observation. Learning the hidden dynamics from indirect observation would be a challenging future work as well.

Another interesting direction for future work would be to use the proposed method for fast and accurate PDE solution by coupling it with traditional numerical solution techniques, for example, pseudospectral methods \cite{rashid2014numerical}, spline collocation methods \cite{abbas2014application} etc. A major challenge for such work would be to guarantee convergence and stability of the solution which is rare among deep learning based PDE solving methods.